\documentclass[journal]{IEEEtran}
\usepackage{amsmath,amsfonts}
\usepackage{algorithmic}
\usepackage{algorithm}
\usepackage{array}
\usepackage[caption=false,font=normalsize,labelfont=sf,textfont=sf]{subfig}
\usepackage{graphicx}
\usepackage{textcomp}
\usepackage{stfloats}
\usepackage{url}
\usepackage{autobreak}
\usepackage{verbatim}
\usepackage{graphicx}
\usepackage{verbatim}
\usepackage{tikz}
\usepackage{threeparttable} 
\usepackage{amsmath}

\usepackage{booktabs}
\usetikzlibrary{positioning}
\usetikzlibrary{shapes,arrows}
\usepackage{tabularx,ragged2e,siunitx}
\usepackage{multirow}	
\usepackage{color, colortbl}
\usepackage{hyperref}
\usepackage{makecell}
\usepackage[capitalise,noabbrev]{cleveref}
\Crefname{figure}{\text{Fig.}}{\text{Figs.}}	
\Crefname{equation}{\text{Eq.}}{\text{Eqs.}}	
\definecolor{Lightgray}{gray}{0.9}
\usepackage{url}

\newcolumntype{L}[1]{>{\raggedright\let\newline\\\arraybackslash\hspace{0pt}}m{#1}}
\newcolumntype{Z}{>{\raggedright\let\newline\\\arraybackslash\hspace{0pt}}X}

\begin{document}

\title{HAD-Gen: Human-like and Diverse Driving Behavior Modeling for Controllable Scenario Generation}

\author{Cheng Wang, Lingxin Kong, Massimiliano Tamborski, Stefano V. Albrecht
\thanks{This work was funded by UK Research and Innovation (UKRI) under the UK government’s Horizon Europe funding guarantee [grant number EP/Z533464/1]. (Corresponding author: \textit{Cheng Wang}).}
\thanks{Cheng Wang is with the School of Engineering and Physical Sciences, Heriot-Watt University, Edinburgh EH14 4AS, United Kingdom (e-mail: cheng.wang@hw.ac.uk).

Lingxin Kong is with the School of Automation and Software Engineering, Shanxi University, Taiyuan 030031, China (e-mail: 202223601006@email.sxu.edu.cn).

Massimiliano Tamborski and Stefano V. Albrecht are with the School of Informatics, University of Edinburgh, Edinburgh EH8 9AB, United Kingdom (e-mail: m.tamborski@sms.ed.ac.uk; s.albrecht@ed.ac.uk).}}

\markboth{}%
{Shell \MakeLowercase{\textit{et al.}}: Driver behavior modeling using offline RL}

\maketitle

\begin{abstract}
Simulation-based testing has emerged as an essential tool for verifying and validating autonomous vehicles (AVs). However, contemporary methodologies, such as deterministic and imitation learning-based driver models, struggle to capture the variability of human-like driving behavior. Given these challenges, we propose HAD-Gen, a general framework for realistic traffic scenario generation that simulates diverse human-like driving behaviors. The framework first clusters the vehicle trajectory data into different driving styles according to safety features. It then employs maximum entropy inverse reinforcement learning on each of the clusters to learn the reward function corresponding to each driving style. Using these reward functions, the method integrates offline reinforcement learning pre-training and multi-agent reinforcement learning algorithms to obtain general and robust driving policies. Multi-perspective simulation results show that our proposed scenario generation framework can simulate diverse, human-like driving behaviors with strong generalization capability. The proposed framework achieves a 90.96\% goal-reaching rate, an off-road rate of 2.08\%, and a collision rate of 6.91\% in the generalization test, outperforming prior approaches by over 20\% in goal-reaching performance. The source code is released at https://github.com/RoboSafe-Lab/Sim4AD.

\end{abstract}


\section{Introduction} \label{introduction}
\IEEEPARstart{A}{utonomous} vehicles (AVs) represent a groundbreaking advancement in transportation technology, with the potential to enhance road safety, reduce traffic congestion, and improve overall mobility efficiency \cite{ma2020artificial}. Over the past decade, significant progress has been made in transforming AVs from theoretical concepts into practical implementations, with numerous prototypes successfully navigating urban environments \cite{albrecht2022aic}. Despite these advancements, ensuring the safety and reliability of AVs across diverse real-world scenarios remains a significant challenge \cite{wang2021online}. 

\begin{figure}[ht]
    \centering
    \includegraphics[width=\linewidth]{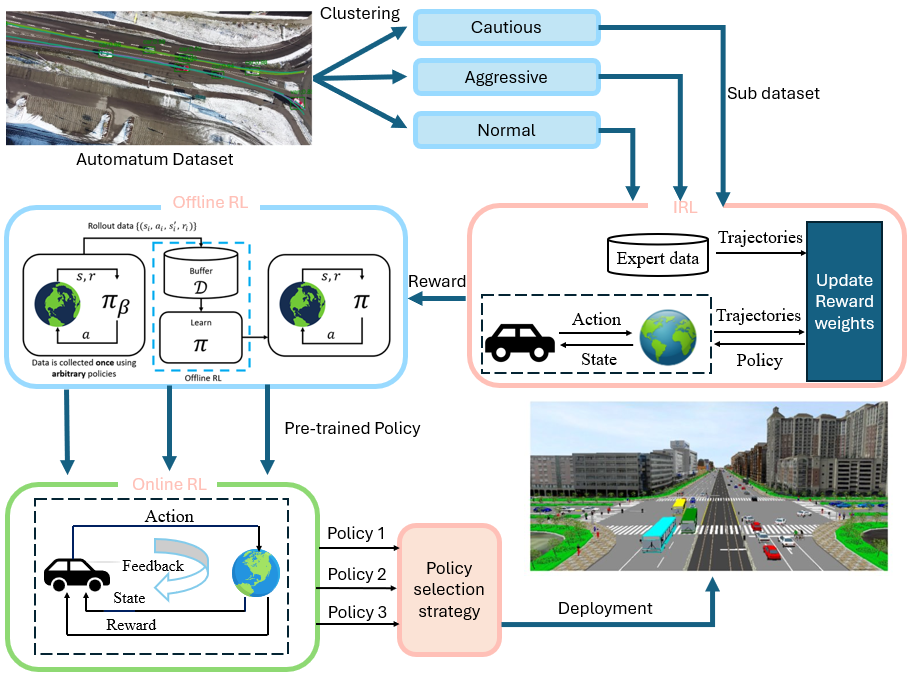}
    \caption{The overall working flow of our proposed method. The driving behavior in a dataset is clustered into distinct driving styles. Each sub-dataset is then used to reconstruct a reward function via IRL. The reconstructed rewards are fundamental for offline RL and MARL, which generate a driving policy for each cluster. The various driving policies are deployed in a simulation based on a policy selection strategy.}
    \label{fig:workflow}
\end{figure}

Simulation-based testing methods are indispensable for addressing this challenge, as field testing is both time-prohibitive and cost-intensive. While these simulations provide a controlled, cost-effective environment for testing various scenarios, in which AVs can perform complex tasks, they often fail to capture the complexity of human driving behavior \cite{wang2023application}. These shortcomings limit their effectiveness in creating realistic traffic environments, which are crucial for thorough AV testing and safety validation. Thus, there is a pressing need for advanced simulation techniques that can accurately model diverse and human-like driving behaviors.

Building on this need, current approaches, such as deterministic driver models (e.g., Intelligent Driver Model, IDM \cite{albeaik2022limitations}) lack the flexibility to adapt to diverse driving scenarios, while imitation learning (IL)-based models (e.g., Behavior Cloning \cite{wang2022high}) struggle with generalization to unseen situations and are prone to compounding errors \cite{booher2024cimrl}. These issues prevent the development of realistic traffic environments, which are crucial for reliable AV testing. Consequently, inverse reinforcement learning (IRL) \cite{sackmann2022modeling} is adopted to reconstruct the reward function of human drivers, which is supposed to be more generalizable than IL as the underlying motivation is learned instead of purely mimicking the driving data. Nevertheless, IRL tends to learn an average driving policy across various driving behaviors, potentially sacrificing behavioral diversity~\cite{ozkan2021inverse}.

He et al. \cite{he2022robust} used RL to model driving behavior. By continually learning from interactions with the environment, RL-based models have the advantage of adapting to unseen situations. Unlike traditional supervised learning, which relies on labeled data, RL agents improve their policies through trial and error, guided by rewards. To meet the requirement of human-like driving, the meticulous design of the reward functions that encode how people drive and behave in traffic interactions is essential \cite{arora2021survey}. However, it is unclear what reward function corresponds to human driving without prior human driving behavior classification \cite{knox2023reward}. 

To this end, we introduce \textbf{H}uman-like \textbf{A}nd \textbf{D}iverse agent behavior modeling for controllable scenario \textbf{Gen}eration (HAD-Gen), a framework for generating diverse, human-like, realistic scenarios, as illustrated in \cref{fig:workflow}, while maintaining model generalizability and controllability in the scenario generation process. HAD-Gen first clusters naturalistic driver behavior to identify distinct driving patterns and then uses IRL to infer reward functions for each cluster. This ensures the preservation of diverse driving styles while maintaining human-like behavior through the implicitly captured objectives and preferences expressed by the inferred reward functions. Subsequently, we leverage offline RL to pre-train the driving policy based on each inferred reward structure to avoid potential bias and unstable training introduced in online RL. Finally, we train multi-agent RL (MARL) \cite{marl-book} policies, resulting in more robust and generalizable simulation environments for AV testing. Depending on which cluster of driving policy is deployed, distinct driving behaviors can be achieved, facilitating the controllability of the test scenarios. 

\textbf{Contributions.} Our research makes the following contributions to the field of AV scenario generation:

1) We propose a new framework called HAD-Gen for scenario generation by combining IRL, offline RL and MARL under each clustered driving pattern to fully leverage the strengths of each algorithm;

2) We establish the theory of HAD-Gen, innovatively leveraging IRL to reconstruct reward functions and employing centralized training with decentralized execution (CTDE) for robust driving policy training;

4) We conduct a comprehensive evaluation of the proposed HAD-GEN framework from multiple perspectives based on the Automatum \cite{spannaus2021automatum} dataset. The results indicate that our method can generate diverse and human-like traffic scenarios while maintaining robust generalization across different conditions.

The rest of this paper is organized as follows. \cref{sec:related_work} introduces related work in the field of driving behavior classification and modeling. In \cref{sec: methodology}, we introduce the proposed method in detail. Extensive experiments in \cref{sec:experiments} demonstrate the effectiveness of the proposed method. \cref{sec:conclusion} concludes the paper and outlines our future work.

\section{Related work}\label{sec:related_work}
In this section, we first discuss driving behavior modeling methods. Then, we present scenario generation methods and discuss relevant simulators.

\subsection{Driving behavior modeling}
Driving behavior modeling methods can generally be divided into three main categories: \textit{heuristic-based methods}, \textit{IL-based methods} and \textit{RL-based methods}. In this section, we provide an overview of the three categories of driving behavior modeling methods and their key characteristics.

\paragraph*{\textbf{Heuristic-based methods}} these methods rely on rule-driven and physics-based models \cite{abuali2016driver}. Kesting et al. \cite{kesting2010enhanced} improved the response mechanism of IDM for handling sudden lane changes and small vehicle gaps, making the proposed model more suitable for implementing adaptive cruise control (ACC) in real vehicles. Shah et al. \cite{shah2023modified} proposed an improved Gipps car-following model to eliminate unrealistic collisions in traffic simulation tools. However, these methods are primarily suited for relatively simple driving behavior simulations, such as car-following, and thus struggle to capture the complexity of real-world traffic scenarios. 

\paragraph*{\textbf{IL-based methods}} these methods directly imitate or copy the decisions and actions of human drivers by learning from expert driving examples. For instance, Zong et al. \cite{zong2023traffic} employed long short-term memory (LSTM) in IL to help the model better mimic the dynamic, time-dependent demonstrations. Zhang et al. \cite{zhang2022systematic} proposed D2Sim, a data-driven trajectory generation model that leverages adversarial learning to effectively capture and generalize complex human driving behaviors from empirical data. 
Unlike behavior cloning, IRL learns the expert’s intent via a reward function instead of direct imitation. 
In \cite{huang2021driving}, maximum entropy (MaxEnt) IRL is employed to infer the reward function of human driving behaviors, which is then used to score the candidate trajectories generated by a polynomial function. Naumann et al. \cite{naumann2020analyzing} utilized MaxEnt IRL to learn the weights of trajectory features and demonstrated how human driving preferences vary significantly under different driving scenarios and behaviors. 
Similarly, Zhao et al. \cite{zhao2022personalized} and Wang et al. \cite{wang2023modeling} employ MaxEnt IRL to learn individual driver preferences from historical data. However, IRL alone cannot capture the full diversity of driving behaviors and may struggle with generalization. We address these limitations by combining IRL with MARL to learn specific policies tailored to each cluster of driving behavior.

\paragraph*{\textbf{RL-based methods}} in these methods the agent interacts with the environment and refines its behavior by maximizing the cumulative reward received. A modeling framework that integrates deep RL with hierarchical game theory was proposed to predict driver behavior in highway scenarios \cite{albaba2021driver}. Methods integrating Monte Carlo Tree Search with Bayesian goal recognition can achieve effective interactive driving, including under occlusions \cite{albrecht2020igp2,hanna2021interpretable,brewitt2023ogrit}. An Attention-based Twin Delayed Deep Deterministic Policy Gradient (ATD3) \cite{fu2020drivers} was introduced to model the driver's temporal attention allocation. Their experiments on real-world vehicle trajectory datasets revealed distinct attention patterns under varying driving conditions. Cornelisse et al. \cite{cornelisse2024human} proposed Human-Regularized PPO (HR-PPO), a MARL algorithm that penalizes deviations from a human reference policy during training to promote realistic driving behavior. Although RL-based methods show strong adaptability and learning capabilities, they heavily rely on manually designed reward functions. If the reward function is not designed properly or fails to account for the diversity of driving behaviors, it may result in poor performance or limited generalization. Consequently, our approach utilizes IRL to infer reward functions for each identified driving style to improve the personalization and generalization of the learned policies.

\subsection{Scenario generation}
Scenario generation for testing AVs primarily consists of two categories: \textit{nominal scenarios generation} and \textit{safety-critical scenarios generation}.

\paragraph*{\textbf{Nominal scenarios}} these scenarios are primarily used to simulate everyday driving behaviors, such as lane keeping, car following, and discretionary lane changes. Kesting et al. \cite{kesting2007general} proposed a general lane-changing model minimizing overall braking induced by lane change (MOBIL), which is widely used in nominal scenarios to simulate naturalistic lane-changing behavior. To improve the realism of nominal scenario generation, Sharath
 et al. \cite{sharath2020enhanced} proposed an enhanced IDM that extends conventional IDM to incorporate lateral decision-making and discretionary lane-changing behavior.

\paragraph*{\textbf{Safety-critical scenarios}} these scenarios involve high-risk or unexpected events, which are infrequent in real-world data. Therefore, the generation of such scenarios is crucial for the testing and validation of autonomous driving systems. Existing generation methods can be categorized into three types: \textit{data-driven generation}, \textit{adversarial generation}, and \textit{knowledge-based generation} \cite{ding2023survey}. Data-driven generation relies on real-world traffic data to generate testing scenarios. An example is the SceneGen proposed by Tan et al. \cite{tan2021scenegen}, which aims to generate realistic traffic scenes by learning from large-scale real-world datasets. Adversarial generation uses techniques like generative adversarial networks to create safety-critical scenarios. For instance, Feng et al. \cite{feng2021intelligent} proposed training background vehicles to perform adversarial maneuvers in a naturalistic driving environment. Finally, knowledge-based generation leverages expert knowledge and predefined rules to construct scenarios. For example, Ding et al. \cite{ding2021semantically} proposed a knowledge-based scenario generation method that integrates semantic traffic rules and object relationships into a tree-structured variational autoencoder to generate adversarial driving scenarios. 

Unlike previous works that focus solely on nominal or safety-critical scenarios, we train a driving policy for each clustered driving style, enabling customized and controllable traffic scenario generation. For instance, by increasing the penetration of deployed aggressive driving policies, we can generate traffic scenarios with more aggressive behavior, leading to potentially more safety-critical scenarios. Conversely, by mixing the policies of different driving styles, we would generate a nominal scenario.

\subsection{Simulators}
High-fidelity simulators play a crucial role in the testing and validation of autonomous vehicles, and several widely used simulators have been developed for this purpose. SUMO \cite{krajzewicz2012recent} and VISSIM \cite{fellendorf2010microscopic} are two representatives for microscopic traffic simulation. CARLA \cite{dosovitskiy2017carla} is another open-source simulator designed specifically for autonomous driving research. It provides diverse simulation assets and supports various autonomous driving approaches, which make it one of the most widely used simulators in the field. LGSVL \cite{rong2020lgsvl} is designed for full-stack testing, enables end-to-end simulation and provides seamless integration with some frameworks such as Autoware \cite{kato2018autoware} and Apollo \cite{xu2019automated}. AirSim \cite{shah2018airsim} has gained significant traction among researchers for its realistic sensor simulation and physics-based vehicle dynamics in AV testing and aerial robotics. Our proposed method can be seamlessly integrated into these simulators, enabling realistic traffic scenarios that accurately reflect diverse, human-like driving behaviors.


\section{Methodology}\label{sec: methodology}
In this section, we introduce the HAD-Gen, including the clustering method, the IRL that we used for inferring the reward functions, the offline RL part based on the derived reward functions for policy initialization, and the MARL training to further optimize the driving policies for interactive driving.

\subsection{Problem statement}
Let $s_t^i$ denote the state of the $i$-th vehicle at time step $t$, which consists of the vehicle's state \{\textit{position, speed, acceleration, heading}\}. The set of states for all $N$ vehicles is represented as
$\mathcal{S}_t = \{ s_t^1, s_t^2, \dots, s_t^N \}$.
The trajectory of the $i$-th vehicle for $T_i$ time steps is defined as $\zeta_i = \left( s_0^i, s_1^i, \dots, s_{T_i - 1}^i \right)$.
We further define a feature extraction function $\phi(\cdot)$ that maps a vehicle's observation to a set of feature vectors. Specifically, for vehicle $i$ at time step $t$, its feature vector $\mathbf{f}_t^i$ is given by $\mathbf{f}_t^i = \phi(o_t^i)$. $o_t^i$ is the partial observation, including ego related state \{\textit{speed, heading, distance to left lane marking, distance to right lane marking}\} and its local observations for each of its surrounding agents on adjacent lanes \{\textit{heading, relative longitudinal distance, relative lateral distance, relative speed, relative acceleration}\}, with a total size of 34.
We aggregate the feature vectors over the vehicle's trajectory and apply a clustering algorithm to classify its driving style $c^i$:
\begin{equation}
c^i = C\Bigl(\sum_{t=1}^{T_i} \mathbf{f}_t^i\Bigr)
\end{equation}
where $c^i \in c_{1, 2, \dots, K}$
represents the driving style cluster for vehicle $i$ with a total of $K$ driving styles. For each $c_{k\in K}$, we aim to learn an optimal policy $\pi_{\rm c}^*$ through RL:
\begin{equation}
\pi_{\rm c}^* = \arg\max_{\pi} \; \mathbb{E}_{\zeta_i \sim \pi} \Biggl[ \sum_{t=0}^{T_i-1} \gamma \, R_{c_k}\bigl(o_t^i, a_t^i\bigr) \Biggr]
\end{equation} 
where $a_t^i$ is the action taken by the $i$-th vehicle at time step $t$, which include \textit{acceleration} and \textit{steering rate}. $\gamma \in[0,1]$ is the discount factor that balances immediate and future rewards, and $R_{c_k}$ is the reward function representing driving style $c_k$, which is inferred through IRL.

\begin{figure*}[ht]
    \centering
    \includegraphics[width=0.9\textwidth]{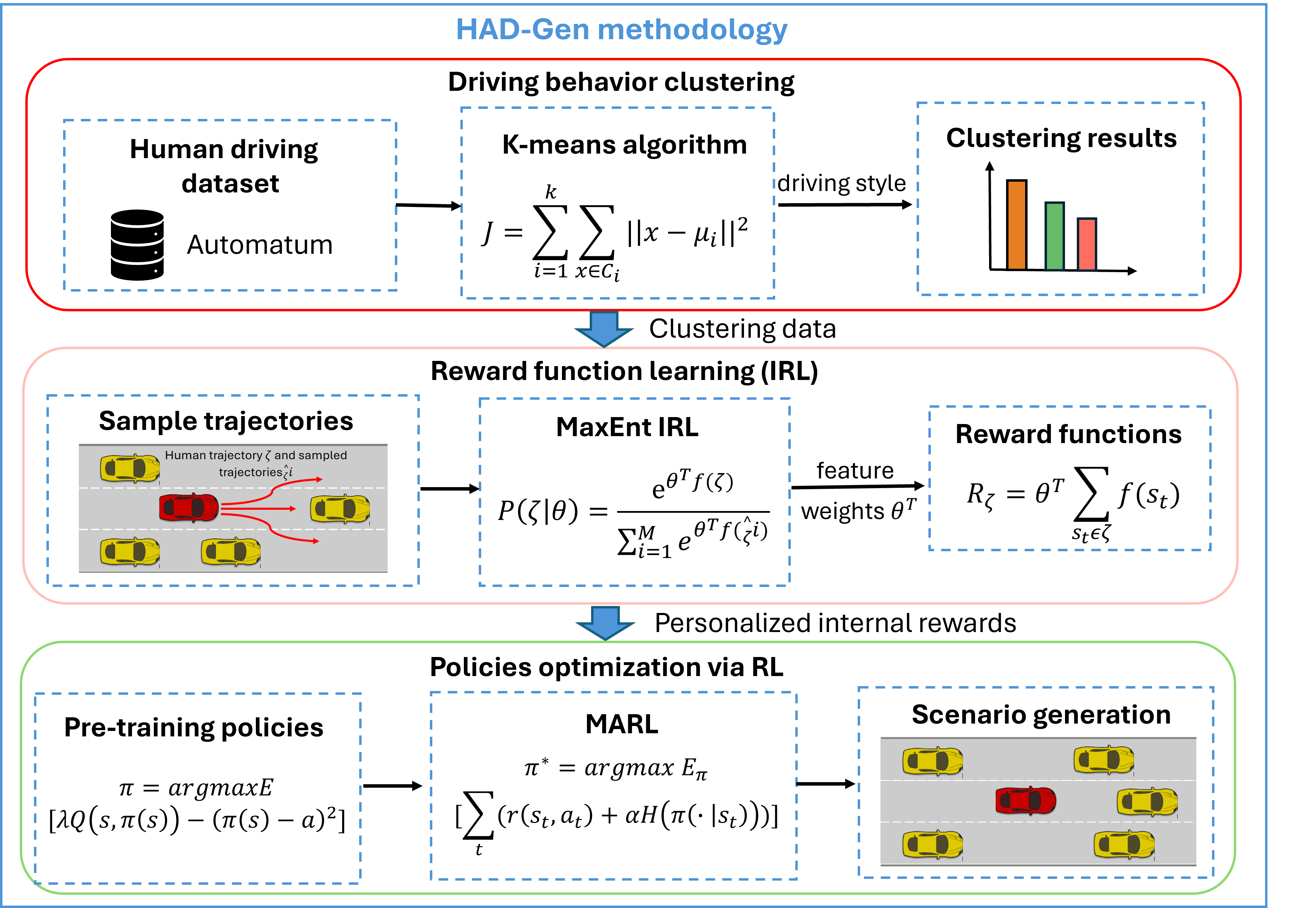}
    \caption{Detailed overview of the proposed HAD-Gen methodology, including driving style recognition, MaxEnt IRL and policy training using MARL.}
    \label{fig:methodology}
\end{figure*}

\subsection{HAD-Gen}
Since drivers typically exhibit diverse driving styles, and a single standard cannot fully capture driving behavior, we first apply an algorithm to categorize driving behavior into different types. This classification helps better represent real-world driving behavior and simulate human traffic scenarios more accurately. Then, based on different types of driving behavior, we use MaxEnt IRL \cite{ziebart2008maximum} to train the weights corresponding to different features and construct a specific style reward function. Finally, we obtain the driving policy for each style through offline RL and online MARL training \cite{marl-book}. An overview of the proposed framework is illustrated in \cref{fig:methodology}.

\subsection{Clustering}
We employ Time-headway (THW) and inverse Time-to-Collision (iTTC) to evaluate the risk associated with each agent's trajectory, as they represent the remaining time to a collision and are defined as:
\begin{equation}
THW = d_{\rm long,rel} / v_{\rm ego}, iTTC = v_{\rm long, rel} / d_{\rm long, rel}
\end{equation}
\( d_{\rm long,rel} \) is the relative distance, \( v_{\rm ego} \) is the vehicle's speed, and \( v_{\rm long, rel} \) is the relative speed to the vehicle ahead.

\begin{figure}[ht]
    \centering
    \includegraphics[width=0.8\linewidth]{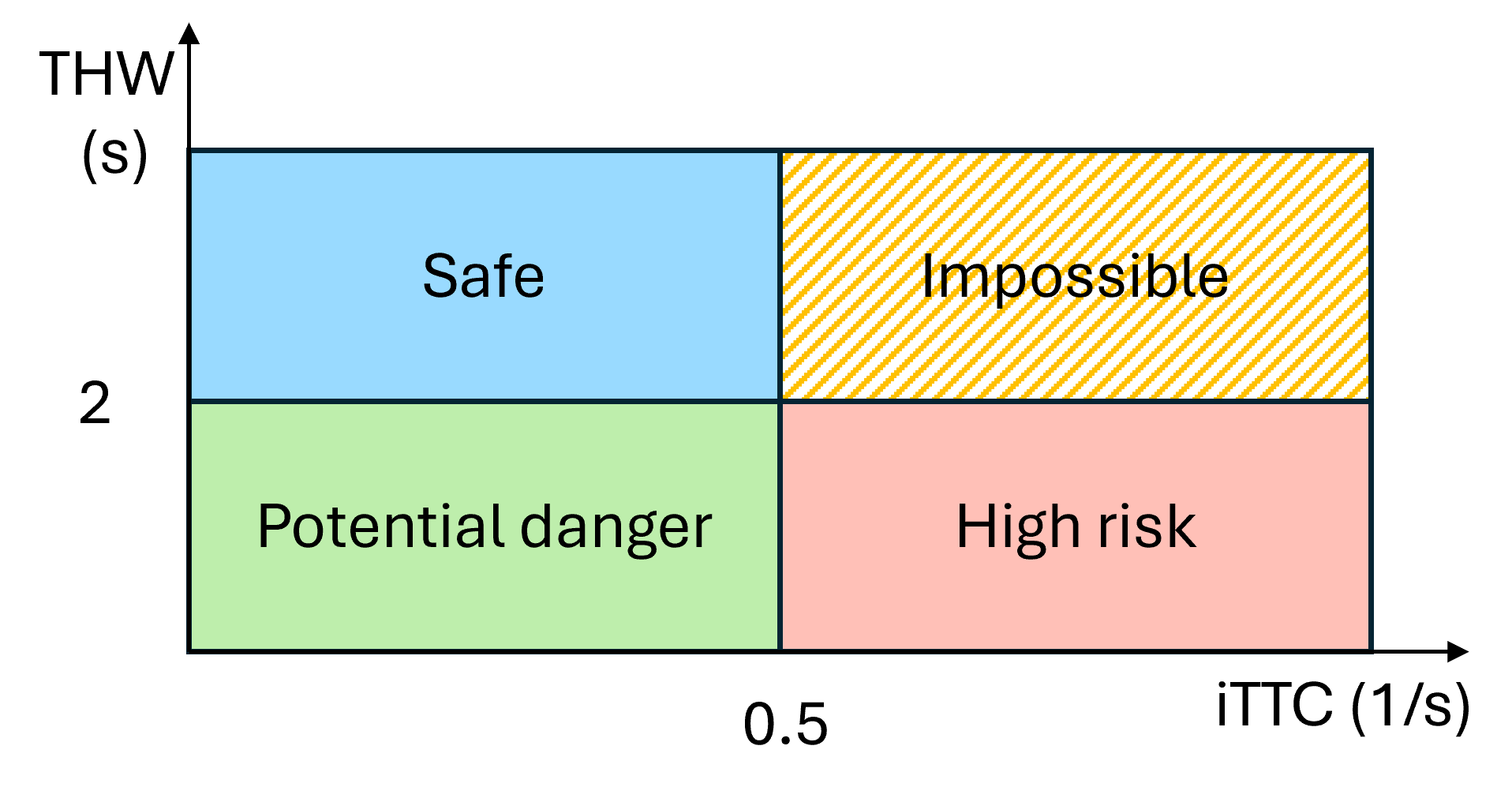}
    \caption{We use iTTC and THW to determine the risk level of a driving behavior. The thresholds are suggested in \cite{xue2019rapid}.}
    \label{fig:clustering_metrics}
\end{figure}

Based on the relative magnitude relationship between THW and iTCC and their thresholds, three risk levels are determined, i.e., \textit{low-risk}, \textit{medium-risk} and \textit{high-risk}, as illustrated in \cref{fig:clustering_metrics}. A trajectory is associated with large iTTC and THW is impossible because $THW \cdot iTTC = v_{\rm long, rel} / v_{\rm ego} \leq 1$. Let $rl_i$ \text{for} $i \in \{1, 2, 3\}$ denote the three risk levels. We segment each agent's trajectory $\zeta$ into three subsegments $Seg_i$ according to their risk levels: 
\begin{equation}
Seg_i = \left\{ S_j(\zeta, rl_i) \mid j \in \{1, 2, \dots, N_i\} \right\}
\end{equation}
where \( S_j(\zeta, rl_i) \) represents the \( j \)-th segment for risk level \( rl_i \), and \( N_i \) is the total number of segments for each risk level \( rl_i \).
Then, we aggregate the segments within each subsegment $Seg_i$ and compute the aggregated segment's length over the total life length $T$ of a vehicle:
\begin{equation}
\text{ratio}_{rl_i} = \sum_{j=1}^{N_i} \ S_j(\zeta, rl_i)  / T
\end{equation}
Finally, the agents are classified using $\text{ratio}_{rl_i}$ as features. Following \cite{martinez2017driving}, we classify the agents into three types: \textit{aggressive}, \textit{normal}, and \textit{cautious}. An \textit{aggressive} driver would have a large proportion of high risk, while a \textit{cautious} driver would have a large proportion of low risk. A \textit{normal} drive in our context is defined as a driver that has a large proportion of medium risk. 

\subsection{Reward reconstruction}
A reward function $r(s_t)$ at a specific state $s_t$ can be defined as:
\begin{equation}
    r(s_t) = \theta^T \textbf{f}(s_t)
\end{equation}
$\textbf{f}(s_t)$ is the extracted feature vector $\textbf{f}(s_t) = [f_1(s_t), f_2(s_t),..., f_k(s_t)]$ that characterizes the state $s_t$. Therefore, the reward of a trajectory $R(\zeta)$ is given as:
\begin{equation}
    R(\zeta) = \sum_t r(s_t) = \sum_t \theta^T \textbf{f}(\zeta) = \theta^T \sum_{s_t\in \zeta} \textbf{f}(s_t)
\end{equation}
$\textbf{f}(\zeta)$ denotes the accumulative features along the trajectory $\zeta$. The problem of IRL is to obtain the reward weights $\theta$ that can generate a driving policy to match the human demonstration trajectories $\mathcal{D} = \{\zeta_1, \zeta_2,..., \zeta_N\}$ consisting of $N$ trajectories.

\paragraph*{\textbf{MaxEnt IRL}} the probability of a trajectory is proportional to the exponential of the reward of that trajectory, given by
\begin{equation}
    P(\zeta | \theta) = \frac{e^{R(\zeta)}}{Z(\theta)} = \frac{e^{\theta^T \textbf{f}(\zeta)}}{Z(\theta)}
\end{equation}
$P(\zeta | \theta)$ is the probability of a trajectory $\zeta$ given reward parameter $\theta$. $Z(\theta)$ is the partition function.

As the partition function $Z(\theta)$ is intractable for continuous and high dimensional spaces, a limited number of trajectories $M$ are generated to approximate the partition function. The probability of a trajectory becomes:
\begin{equation} \label{eq:IRLApproxProb}
    P(\zeta | \theta) \approx \frac{e^{\theta^T \textbf{f}(\zeta)}}{\sum_{i=1}^M e^{\theta^T \textbf{f}(\hat{\zeta}^i)} }
\end{equation}
$\hat{\zeta}^i$ is a generated trajectory $i$ that has the same initial state as $\zeta$, $\textbf{f}(\hat{\zeta}^i)$ is the feature vector of the trajectory.

The goal of maximum entropy IRL is to adjust the weights $\theta$ to maximize the likelihood of expert demonstrations under the trajectory distribution:
\begin{equation} \label{eq:IRLObjective}
    \max_{\theta} \mathcal{J}(\theta) = \max_{\theta} \sum_{\zeta \in \mathcal{D}} \log P(\zeta | \theta)
\end{equation}

Substituting $P(\zeta | \theta)$ in \cref{eq:IRLObjective} with \cref{eq:IRLApproxProb}, the objective function $\mathcal{J}(\theta)$ can be rewriteen as:
\begin{equation}
    \mathcal{J}(\theta) = \sum_{\zeta \in \mathcal{D}} [\theta^T \textbf{f}(\zeta) - \log \sum_{i=1}^M e^{\theta^T \textbf{f}(\hat{\zeta}^i)}]
\end{equation}
The gradient of the objective function $\mathcal{J}(\theta)$ is:
\begin{equation}
    \nabla_\theta \mathcal{J}(\theta) = \sum_{\zeta \in \mathcal{D}} [\textbf{f}(\zeta) -  \sum_{i=1}^M \frac{e^{\theta^T \textbf{f}(\hat{\zeta}^i)}}{\sum_{i=1}^M e^{\theta^T \textbf{f}(\hat{\zeta}^i)}} \textbf{f}(\hat{\zeta}^i)]
\end{equation}
$\textbf{f}(\zeta)$ is the feature vector of a human demonstrated trajectory, $\hat{\zeta}^i$ is one of the generated trajectories that has the same initial state as $\zeta$. The gradient can be seen as the difference of feature expectations between human demonstration trajectories and the generated ones:
\begin{equation}
    \nabla_\theta \mathcal{J}(\theta) = \sum_{\zeta \in \mathcal{D}} [\textbf{f}(\zeta) - \sum_{i=1}^M P(\hat{\zeta}^i | \theta) \textbf{f}(\hat{\zeta}^i)]
\end{equation}

We generate sampled trajectory $\hat{\zeta}^i$ for a vehicle using polynomial-based trajectory planning in Frenet coordinates, which is expressed by:
\begin{equation}
x(t) = b_0 + b_1 t + b_2 t^2 + b_3 t^3 + b_4 t^4
\end{equation}
\begin{equation}
y(t) = a_0 + a_1 t + a_2 t^2 + a_3 t^3 + a_4 t^4 + a_5 t^5
\end{equation}
where $x(t)$, $y(t)$ represents the longitudinal and lateral position at time $t$, respectively. The coefficients \( \mathbf{a} = [a_0, a_1, \dots, a_5]^T \) and \( \mathbf{b} = [b_0, b_1, \dots, b_4]^T \) are determined by soving the boundary equations.

\subsection{Pre-training}
We utilize offline RL to pre-train the policies using the above reconstructed reward functions. Offline RL is characterized by the absence of interaction with the environment to collect new data. This is beneficial if the data collection is expensive, risky or challenging. TD3 \cite{fujimoto2018addressing} is an actor-critic algorithm that improves upon the stability of its predecessor, DDPG (Deep Deterministic Policy Gradient) \cite{lillicrap2015continuous}, by incorporating three key strategies:
\begin{itemize}
    \item Twin Critic Networks: TD3 uses two separate critic networks to mitigate positive bias in the policy improvement step by taking the minimum value of the two critics to reduce overestimation;
    \item Delayed Policy Updates: The policy network is updated less frequently than the critic networks, which helps in decoupling the dependency between policy and value estimates and stabilizes learning;
    \item Policy smoothing: Adding noise to the target policy's actions to prevent the policy from exploiting sharp gradients in the Q-function.
\end{itemize}

However, TD3 like typical offline RL suffers from a distribution shift where the policy tends to poorly estimate the value of state-action pairs not contained in the dataset, resulting in poor performance. To address this problem, the idea of keeping the learned policy close to the behavior policy in the dataset has been proposed, emerging various methods such as BCQ \cite{fujimoto2019off} and CQL \cite{kumar2020conservative}. Due to the complicated hyperparameter tuning process, Fujimoto and Gu \cite{fujimoto2021minimalist} introduced TD3+BC by adding a BC term to regularize the policy with minimal changes to the original TD3 algorithm, as defined as:
\begin{equation}
\pi=\mathop{\arg\max}\limits_{\pi}\mathbb{E}_{(s,a)\sim \mathbf{D}}[\lambda Q(s, \pi(s))-(\pi(s)-a)^{2}]
\end{equation}
where $\lambda$ is a hyperparameter to control the strength of the regularizer and expressed by:
\begin{equation}
    \lambda = \frac{\alpha}{\frac{1}{N} \sum |Q(s_i, a_i)|}
\end{equation}
$N$ is the total transitions in the dataset and $\alpha$ is the weight, which favors BC ($\alpha=1$) or RL ($\alpha=4$). 


\subsection{Online training via MARL}
Online RL involves continuous interaction with the environment during training and directly from the interaction data. In this paper, we employ the Soft Actor-Critic (SAC) \cite{haarnoja2018soft}
algorithm to learn different styles of driving behavior. Unlike other RL algorithms, SAC not only maximizes long-term rewards but also aims to maximize the entropy of the policy, which enhances exploration and introduces randomness during the learning process. Consequently, the optimal policy $\pi^*$ is defined as
\begin{equation}
    \pi^* = \arg\max_{\pi} \mathbb{E}_\pi \left[ \sum_t \Big( r(o_t, a_t) + \alpha H(\pi(\cdot | o_t)) \Big) \right]
\end{equation}
where $r(o_t, a_t)$ denotes the reward value obtained by taking action $a_t$ in observation $o_t$, $\alpha$ is a regularization coefficient that controls the importance of the entropy term, and $H(\pi(\cdot|s_t))$ represents the entropy of the policy $\pi^*$. Its mathematical expression is:
\begin{equation}
    H(\pi(\cdot | s_t)) = -\mathbb{E}_{a_t \sim \pi} \big[\log \pi(a_t | o_t)\big]
\end{equation}
Entropy regularization increases the degree of exploration of the RL algorithm, helps accelerate subsequent policy learning, and reduces the possibility of the policy falling into a poor local optimum.

SAC is an algorithm based on the Actor-Critic architecture. It employs two independent Critic networks along with the corresponding target networks to reduce the overestimation of Q-values. The delayed updates of the target networks further enhance the stability of training. The temporal difference (TD) target expression of the Critic network is
\begin{equation}
y_i = r_i + \gamma \min_{j=1,2} Q_{\omega_j}(o_{i+1}, a_{i+1}) - \alpha \log \pi_\theta(a_{i+1} | o_{i+1})
\end{equation}
where $\gamma$ is the discount factor that determines the importance of long-term cumulative rewards. $Q_{\omega_j}(o_{i+1}, a_{i+1})$ values are computed by the parameters $w_j$ of the two target networks at the time step $t+1$. Furthermore, the loss function of the Critic network is expressed as
\begin{equation}
L = \frac{1}{N} \sum_{i=1}^N \left( y_i - Q_{\omega_j}(o_i, a_i) \right)^2
\end{equation}
where $N$ represents the number of batch samples.
Meanwhile, the above entropy adjustment term introduced in the Actor network balances exploration and exploitation. The loss function of the Actor network is given by
\begin{equation}
L_\pi(\theta) = \frac{1}{N} \sum_{i=1}^N \left( \alpha \log \pi_\theta(\tilde{a}_i | o_i) - \min_{j=1,2} Q_{\omega_j}(o_i, \tilde{a}_i) \right)
\end{equation}
where the action $\tilde{a}_i$ is obtained from the current policy distribution and using the reparameterization trick \cite{kingma2015variational}. In our multi-agent setting, all agents share the same SAC policy via parameter sharing \cite{christianos2021scaling}. Multiple agents make decisions simultaneously in an interactive environment, and their experiences are collected into a shared replay buffer for updating the SAC policy.

\section{Experiments}\label{sec:experiments}
In this section, we provide a detailed description of the dataset utilized in this study, the feature definitions employed for reward function reconstruction, the hyperparameter settings for training, and the specific design of the experiments.

\subsection{Experiment design}
We aim to answer three fundamental research questions regarding scenario generation by designing the following experiments.

\textbf{Q1} How can we test the \textbf{diversity} of the generated traffic scenarios?

Diverse driving behavior is crucial in traffic scenario generation. Diversity in our context means that the generated traffic scenarios exhibit diverse driving styles and thus contribute to diverse maneuvers. We define three sub-experiments, each corresponding to one of three driving polices: \textit{cautious, normal and aggressive} that control all the agents in each episode. We then analyze whether the three sub-experiments show distinct distributions of key scenario parameters. Additionally, we conduct another sub-experiment to assess whether the different driving policies result in diverse driving maneuvers within the same scenario.

\textbf{Q2} How can we test the \textbf{human-likeness} of the generated traffic scenarios?

We deploy our three trained policies while keeping the initial states of all involved agents the same as the initial states in the original dataset and keeping identical driving style proportions to that episode. We compute the distributions of each macroscopic metric defined in \cref{tab:evaluation_metrics} based on generated scenarios and original scenarios, whereas the distribution differences are gauged by Jensen-Shannon Distance (JSD) and Hellinger Distance (HD). JSD is defined as the square root of the JS divergence, which is a symmetric measure computed by introducing the average distribution of two probability distributions and calculating the Kullback-Leibler (KL) divergence relative to this average. This approach avoids numerical instability caused by zero probabilities in the distributions and ensures that the metric satisfies properties such as the triangle inequality. The JSD between the probability distributions $p$ and $q$ is expressed as:
\begin{equation}
m = \frac{1}{2}(p + q)
\end{equation}
\begin{equation}
\mathrm{KL}(p \,\|\, q) 
= \sum_i p_i \log\left(\frac{p_i}{q_i}\right)
\end{equation}
\begin{equation}
\mathrm{JSD}(p \,\|\, q) 
= \sqrt{\frac{1}{2}\mathrm{KL}(p \,\|\, m) 
+ \frac{1}{2}\mathrm{KL}(q \,\|\, m)}
\end{equation}
where $m$ represents the average distribution of $p$ and $q$.
In addition, the HD is a symmetric geometric measure that measures the differences in probability mass between two distributions. It is defined as:

\begin{equation}
H(p, q) 
= \frac{1}{\sqrt{2}} \sqrt{ \sum_{i=1}^n \left(\sqrt{p_i} - \sqrt{q_i}\right)^2 }
\end{equation}

\textbf{Q3} How can we compare the \textbf{generalizability} between HAD-Gen and other baselines?

We use unseen data recorded in the dataset to demonstrate the generalizability of our proposed model. In this experiment, we deploy one of the trained policies with 100\% penetration, which will control all the agents in each episode. We compared our method with the baselines using microscopic metrics defined in \cref{tab:evaluation_metrics}. This evaluation aims to assess the performance of different methods in real-world scenarios to test whether our policies will behave plausibly in scenarios where other baselines fail. These sub-experiments will strongly support the generalization of our model. 

\subsection{Evaluation metrics}
We use four microscopic metrics to evaluate the diversity and human-likeness of driving policies: \textit{Speed}, \textit{Distance}, \textit{THW} and \textit{iTTC} distributions. Speed distribution reflects the vehicle's driving efficiency and adaptability to traffic flow; Distance distribution refers to the minimum distance between the current vehicle and surrounding vehicles; THW and iTTC are important reference indicators for measuring the safety and driving style of AVs.

In the generalization test, we use three key macroscopic metrics to evaluate the effectiveness of different driving policies include \textit{goal-reaching rate}, \textit{off-road rate} and \textit{collision rate}, which are all computed as the ratio of specific events to the total number of agents. Specifically, the goal-reaching rate measures the proportion of agents that successfully traversed the predefined map length and safely exited the map boundary. The off-road rate evaluates the proportion of agents that crossed the outermost lane boundaries (i.e., the leftmost or rightmost road edges). The collision rate represents the ratio of agents involved in collisions with other agents during their trajectories. The following \cref{tab:evaluation_metrics} summarizes these metrics and their respective categories.

\begin{table}[!ht] 
\caption{Metrics to evaluate the driving policy trained by HAD-Gen in both microscopic and macroscopic levels}
\centering
\begin{tabular}{lll}
\toprule
Scope & Category & Metrics  \\ 
\midrule
\multirow{5}{*}{\hspace{-0.7em}\vspace{2.5em} Macroscopic} & & Collision rate \\
& \textit{Generalization} & Goal-reaching rate \\
& & Off-road rate \\
\midrule
\multirow{2}{*}{\hspace{-0.7em}\vspace{-0.3em} Microscopic} & \textit{Diversity} & Speed, Distance, \\
& \textit{Human-likeness} & THW, and iTTC distribution \\
\bottomrule
\end{tabular}
\label{tab:evaluation_metrics}
\end{table}

\subsection{Dataset}
We use the Automatum \cite{spannaus2021automatum} dataset to validate the driving behavior modeling method proposed in this paper. This dataset is collected by drones in highway scenarios and contains 30,134 kilometers of driving data and 30 hours of recording, which can provide comprehensive traffic flow dynamics and driving behavior data. The diverse driving behaviors captured in this dataset are necessary for clustering different driving styles and learning the corresponding driving policies. 

The Automatum dataset contains detailed trajectory data of vehicles, including trajectory length, speed, pose (position and orientation), and sampling interval, etc. The object trajectories are highly accurate, with a validated relative speed error of less than 0.2\%. This study selected 13 episodes from a map in the dataset as experimental data, using the first 11 episodes as the training set and the last 2 episodes as the test set.

\subsection{Implementation}
Here, we define the features used in MaxEntIRL, representing key mappings of critical state variables from the environment. We apply feature normalization using the maximum and minimum values of the calculated features to scale them within a defined range. We also develop a MARL simulator for online training, as demonstrated in our open-sourced code.   

\textit{Efficiency}: We use the absolute speed $v_{\rm ego}$ of the ego vehicle to measure the driving efficiency of the ego vehicle.

\textit{Comfort}: We use the longitudinal $a_{\rm long}$, lateral acceleration $a_{\rm lat}$ and longitudinal jerk $J_{\rm long}$ of the ego vehicle to measure the motivation of driving comfort of human drivers. Additionally, we consider the distance to the lane centerline $d_{\rm c}$ and its rate of change $\dot{d}_{\rm c}$, as frequent lane changes or continuous corrections can negatively affect driving stability and passenger comfort. 

\textit{Safety}: THW is a typical critical metric to quantify the safety level of driving. We compute both $THW_{\rm f}$ and $THW_{\rm r}$ in terms of the ego vehicle, representing the time headway of the leading and following vehicles, respectively. In addition, we use two features $avail_{\rm l}$ and $avail_{\rm r}$ to determine whether the left and right lanes are available. 

\paragraph*{\textbf{Baselines}} We use behavioral cloning (BC \cite{zong2023traffic}), offline RL (TD3+BC \cite{fujimoto2021minimalist}), and online RL (SAC \cite{haarnoja2018soft}) as the baselines. BC relies on high-quality expert data to learn the policy, while TD3+BC learns from historical experience without interacting with the environment. SAC relies on the rewards gained from interacting with the environment for training. These baselines are selected to provide a comprehensive comparison across different paradigms of RL.
The hyperparameters used for training the BC, TD3+BC and SAC algorithms are shown in \cref{tab:hyperparameters}.
\begin{table}[h!]
\centering
\caption{Key Hyperparameters for Different Algorithms}
\begin{tabular}{lccc}
\toprule
Parameter & BC &TD3+BC & SAC\\ 
\midrule
Learning rate                & 0.001 & $3e-4$ & 0.001 \\
Batch size                   & 128& 128  & 256 \\
Discount factor ($\gamma$)   & --- & 0.99& 0.99 \\
Target update interval       & ---  & 1 & 1 \\
Policy update frequency      & --- & 2& 2 \\
Replay buffer size           & --- & $10^6$& $10^5$ \\
Target smoothing coefficient ($\tau$) &--- &0.005 &0.005 \\
Entropy coefficient ($\alpha$) & --- & --- & 0.2 \\
\bottomrule
\end{tabular}
\label{tab:hyperparameters}
\end{table}

\section{Results analysis}\label{sec:Results analysis}
In this section, we present the clustering and IRL results. Following this, we provide answers to the three research questions, drawing on the experimental results and our in-depth analysis.

\subsection{Clustering and IRL results}
We use the silhouette value (the $\uparrow$, the better), Davies-Bouldin Index (the $\downarrow$, the better), and Calinski-Harabasz Index (the $\uparrow$, the better) \cite{gagolewski2021cluster} to evaluate the effects of three classification methods: $k$-means, hierarchical clustering (HC) \cite{murtagh2012algorithms}, and Gaussian Mixture Model (GMM) \cite{reynolds2009gaussian}. $k$-means obtained the best clustering result, which has a silhouette value of 0.83 (maximum of the three), the Davies-Bouldin Index of 1.23 (minimum of the three), and the Calinski-Harabasz Index of 69,827.37 (maximum of the three). By analyzing the centroid of each cluster, we can identify the most influential feature, which reveals the dominant risk level $rl_i$ within the cluster. According to the principle that a cautious driver exhibits a low high-risk proportion, while an aggressive driver has a high high-risk proportion, with a normal driver falling in between, we can classify the driving style of each cluster accordingly. The final classification results include 7776 vehicles in the normal category, 2297 in the cautious category, and 7 in the aggressive category.
\begin{table}[h!]
\centering
\caption{Weights of different features obtained from IRL training}
\setlength{\tabcolsep}{2pt}
\resizebox{\columnwidth}{!}{
\begin{tabular}{lcccccccccc}
\toprule
Feature & $v_{\rm ego}$ & $a_{\rm long}$ & $a_{\rm lat}$ & $J_{\rm long}$ & $THW_{\rm f}$ & $THW_{\rm r}$ & $d_{\rm c}$ & $\dot{d}_{\rm c}$ & $avail_{\rm l}$ & $avail_{\rm r}$ \\
\midrule
\vspace{0.3em}Normal & -1.36 & -0.64 & 4.49 & -3.28 & -0.52 & 0.29 & 1.38 & -2.80 & -0.12 & 0.45 \\
\vspace{0.3em}Aggressive & -4.41 & -2.85 & 10.10 & 6.19 & -1.30 & 1.58 & 7.46 & -8.43 & 0.14 & 2.84 \\
\vspace{0.3em}Cautious & -2.86 & -1.18 & 2.90 & -2.82 & 0.17 & -0.02 & 1.98 & -3.75 & 0.13 & 0.26 \\
\bottomrule
\end{tabular}
}
\label{tab:weights}
\end{table}

We infer the reward function under each classification using IRL. \cref{tab:weights} shows the inferred feature weights for the three classifications. There are significant differences in the parameters of different styles, which demonstrates that our modeling approach can reflect the diversity of driving behaviors. Further, the features $a_{\rm lat}$ and $d_{\rm c}$ for aggressive drivers have large weights, indicating that these drivers tend to change lanes abruptly and drive off the centerline of a lane. Also, the $THW_{\rm f}$ to the vehicle in front is a negative maximum in this category, which further validates that this policy accepts higher driving risks. In contrast, the cautious policy has the smallest $a_{\rm lat}$ and a positive $THW_{\rm f}$, suggesting that this policy may favor conservative behaviors of keeping a larger distance from the vehicle in front and overtaking less often. The normal policy falls in between. Meanwhile, both normal and cautious policies tend to favor smaller $v_{\rm ego}$, $a_{\rm long}$, $J_{\rm long}$, and $\dot{d}_{\rm c}$, reflecting that drivers consider both safety and passenger comfort. Whereas the larger $J_{\rm long}$ in the aggressive policy ignores passenger comfort.
\begin{figure}[htbp]
    \centering

    \subfloat[\scriptsize Speed distribution]{%
        \includegraphics[width=0.23\textwidth]{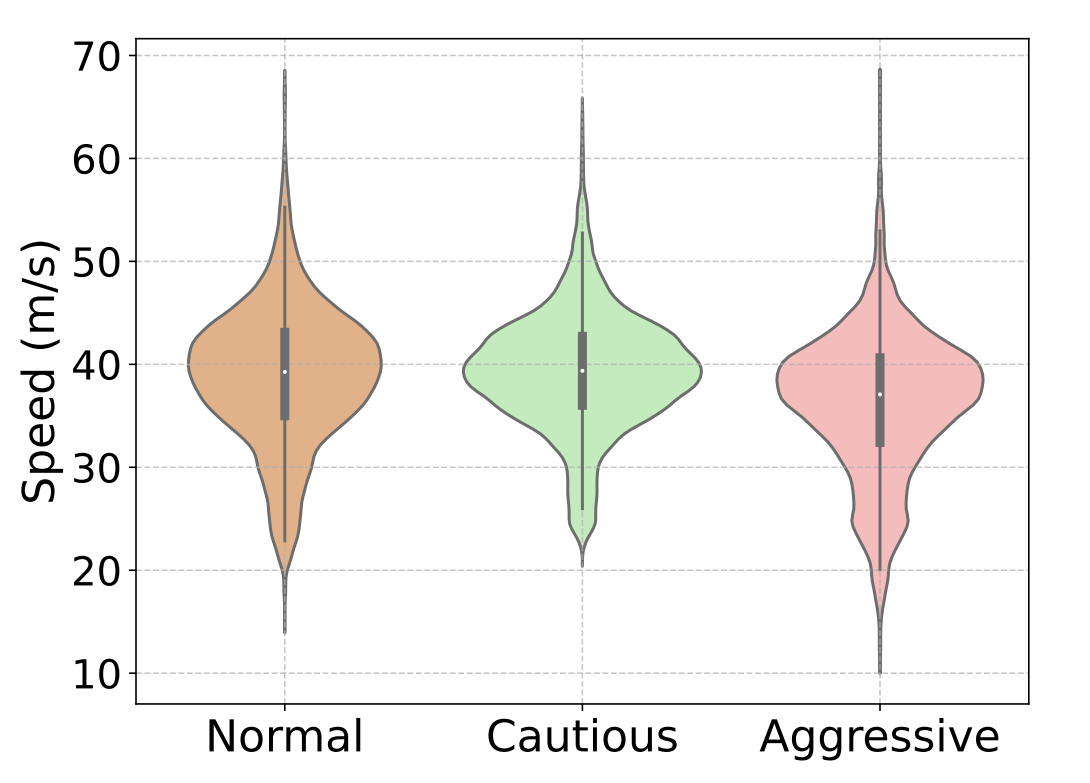}%
        \label{fig:speed}%
    }
    \hspace{0.005\textwidth} 
    \subfloat[\scriptsize  Distance distribution]{%
        \includegraphics[width=0.23\textwidth]{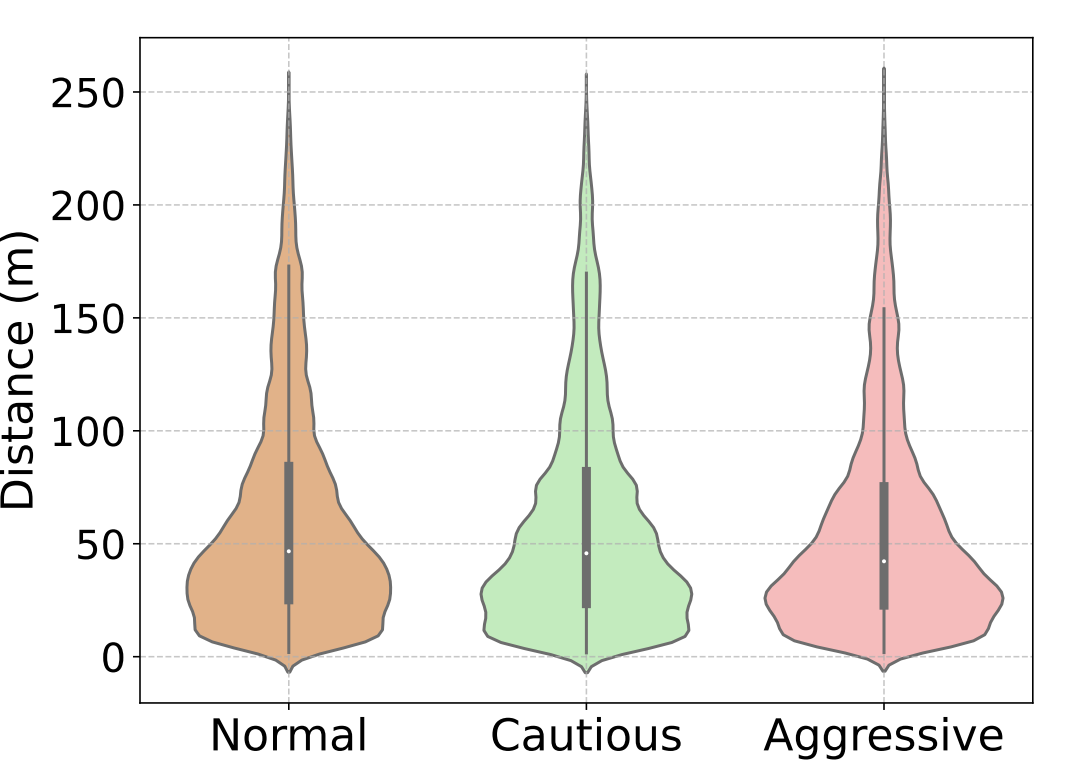}%
        \label{fig:distance}%
    }

    \vspace{0.0cm} 

    \subfloat[\scriptsize  THW distribution]{%
        \includegraphics[width=0.235\textwidth]{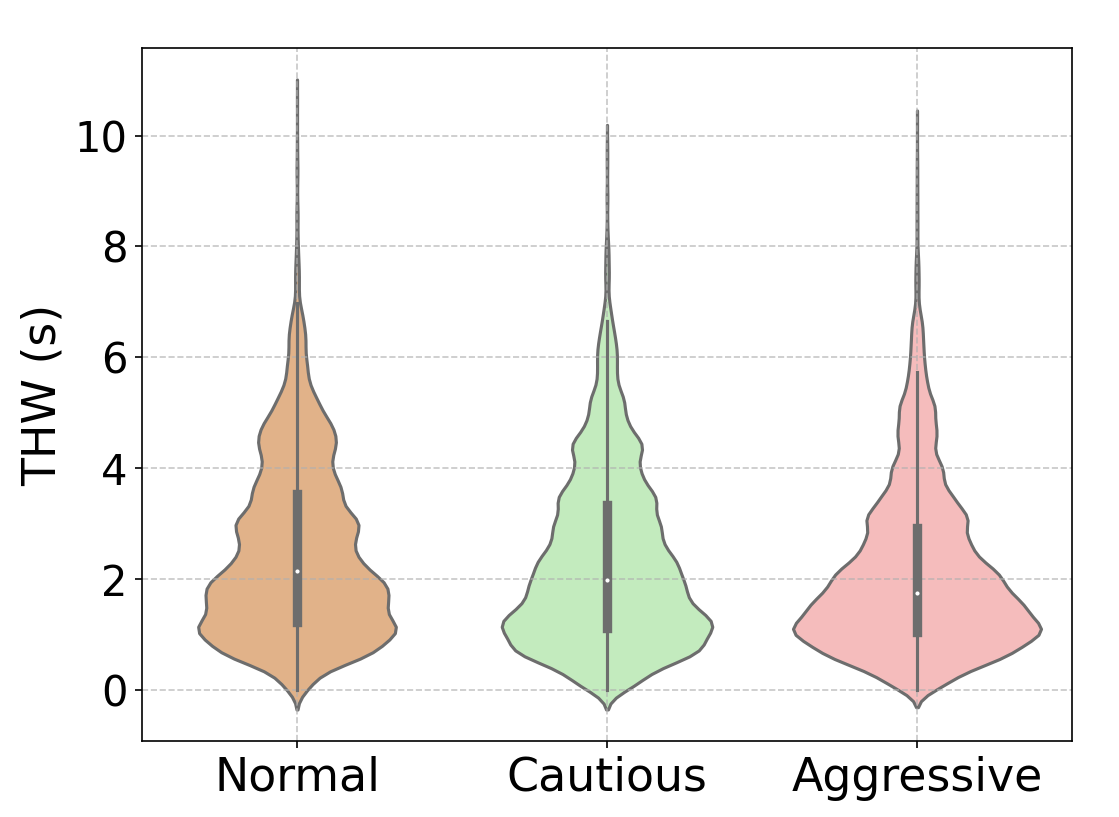}%
        \label{fig:thw}%
    }
    \hspace{0.005\textwidth}
    \subfloat[\scriptsize  iTTC distribution]{%
        \includegraphics[width=0.225\textwidth]{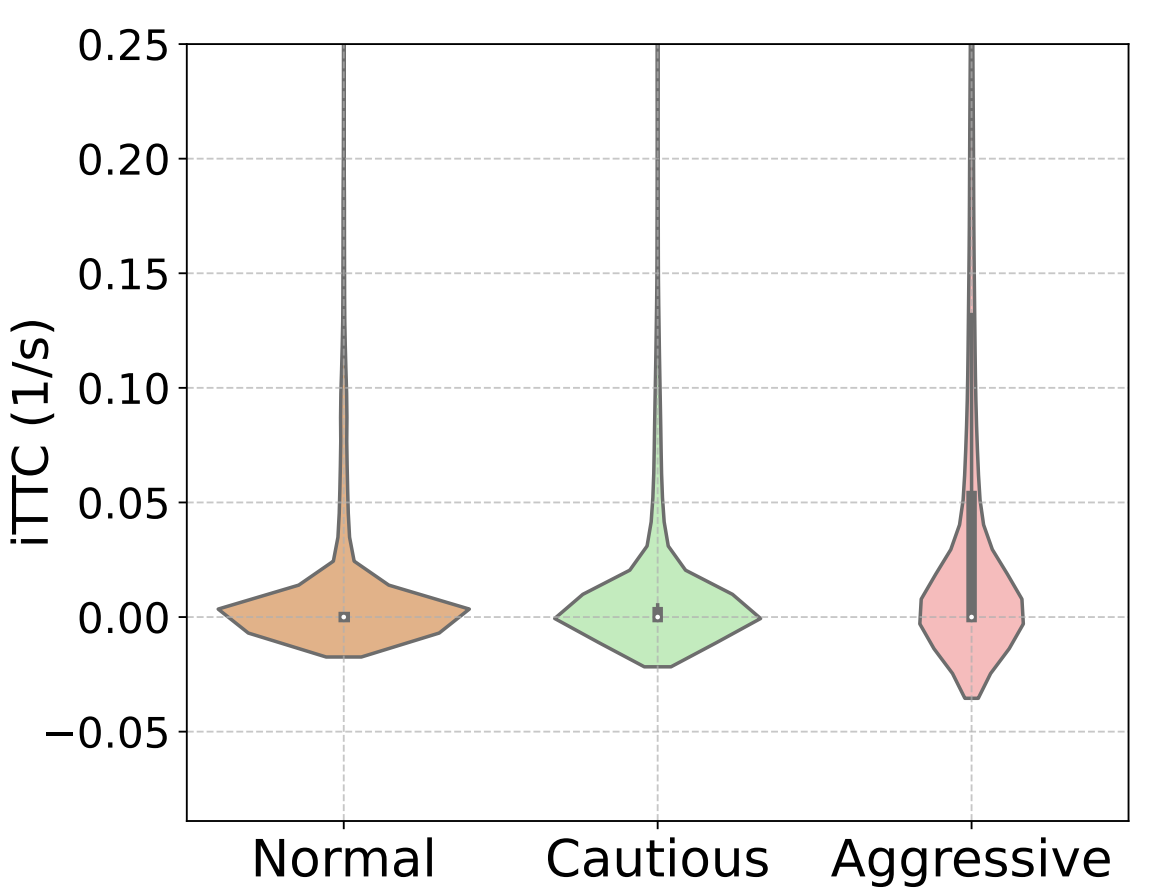}%
        \label{fig:ittc}%
    }

    \caption{Distribution of key metrics demonstrating the diversity of generated scenarios for different driving policies.}
    \label{fig:diversity}
\end{figure}

\begin{figure}[htbp]
    \centering
    \includegraphics[width=\columnwidth]{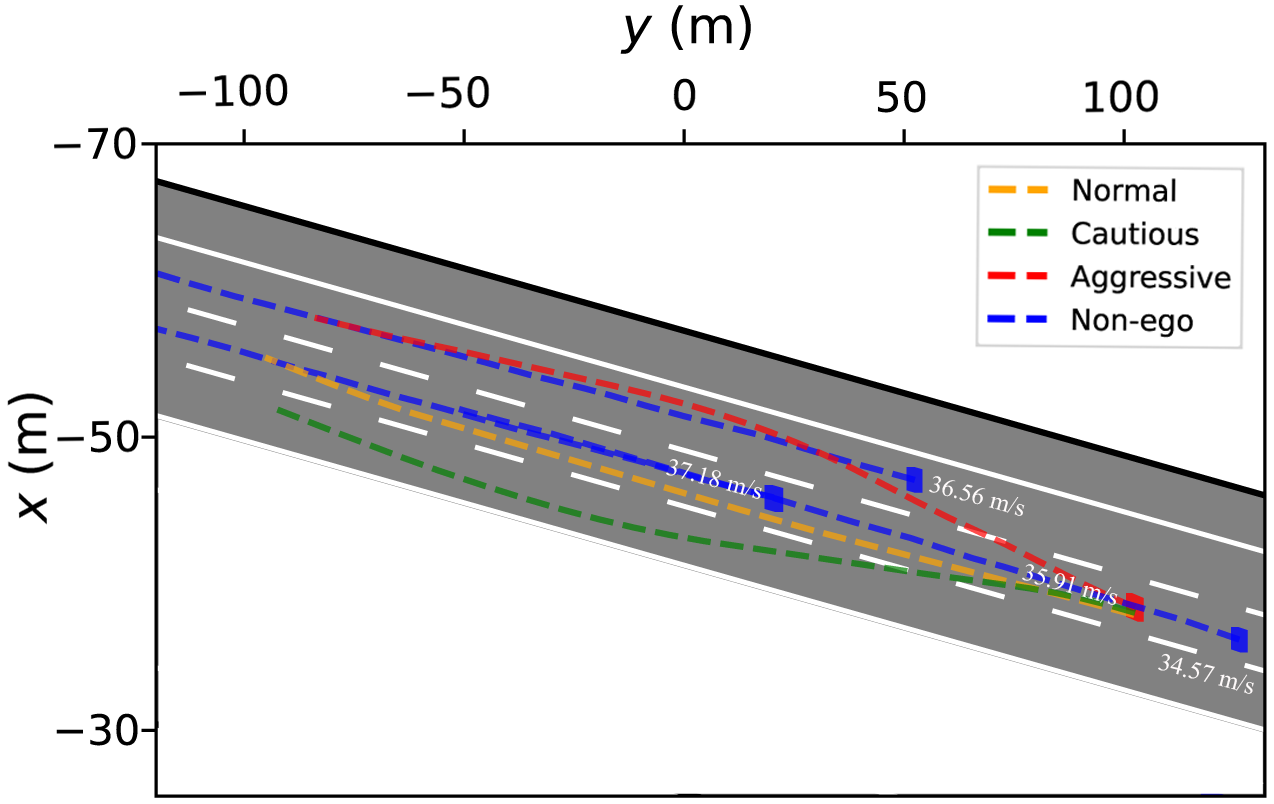} 
    \caption{The driving policies corresponding to different driving styles show different driving behaviors in the same scenario.}
    \label{fig:trajectory_visualization}
\end{figure}

\subsection{Agent behavior modeling results }
\textbf{Q1-Diversity} 
The violin plots in \cref{fig:diversity} illustrate the diversity of scenarios generated by different driving policies. The speed distribution in \cref{fig:speed} for the cautious policy is the most concentrated, with a median around 40 m/s and the lowest dispersion. In contrast, the speed distribution for the aggressive policy is the widest, with the highest dispersion, and includes extreme values not observed in the other two strategies. This might be due to the agent exhibiting high-risk behaviors, such as driving too close to the front vehicle or accelerating quickly to reduce the gap. In comparison, the normal policy exhibits a moderate distribution difference, falling between the first two.
\cref{fig:distance} shows that the cautious and normal policies have a median minimum distance of about 50 meters, while the aggressive policy has a slightly lower median and higher density of distances between 0-50 meters, indicating closer vehicle spacing.

In \cref{fig:thw}, the aggressive policy generates scenarios with the least THW dispersion, a higher density in the 0-2s range, and a median value less than 2s. This suggests that the aggressive policy tends to generate high-risk driving scenarios. In contrast, the cautious and normal policies have a median around 2s and a wider distribution. In particular, the cautious policy has the smallest extremes, aligning with its stable behavior and concentrated speed distribution.
Finally, \cref{fig:ittc} shows that the cautious and normal policies have concentrated iTTC distributions near 0, indicating low-risk scenarios, while the aggressive policy has a wider distribution with a higher density in areas of larger iTTC, suggesting higher risk.

We further use \cref{fig:trajectory_visualization} to demonstrate that different driving policies exhibit different maneuvers even in the same scenario. The different maneuvers further indicate that the HAD-Gen framework successfully captures diverse driving behaviors within different styles through a combination of driving style recognition and IRL to construct reward functions. By deploying driving strategies of different styles and adjusting the ratio between them, we can systematically generate a range of traffic scenarios, spanning from homogeneous to heterogeneous behaviors and from low-risk to high-risk behaviors. 

\begin{figure}[htbp]
    \centering
    \subfloat[\scriptsize Speed (m/s)]{%
        \includegraphics[width=0.235\textwidth]{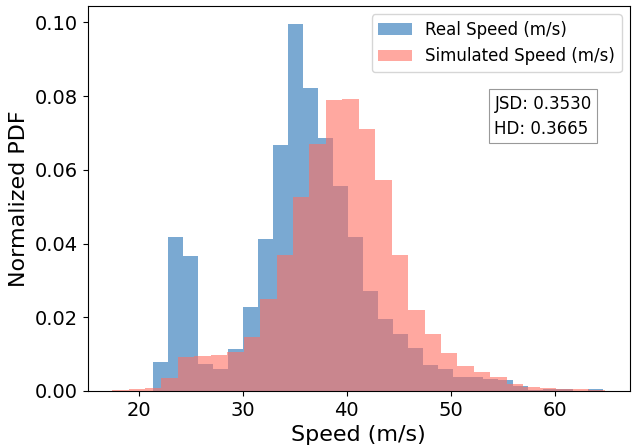}
        \label{fig:speed1}%
    }
    \hspace{-0.00\textwidth} 
    \subfloat[\scriptsize Distance (m)]{%
        \includegraphics[width=0.235\textwidth]{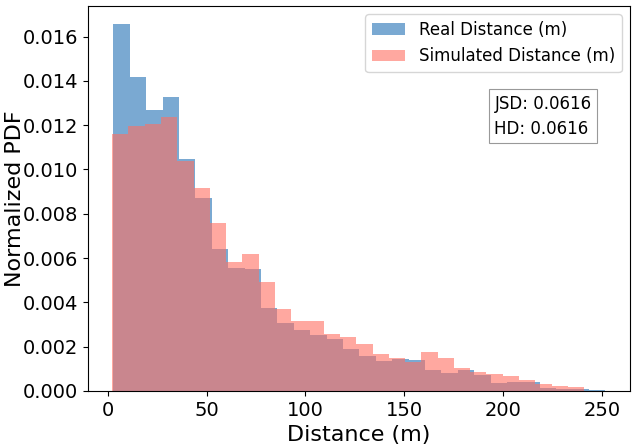}%
        \label{fig:distance2}%
    }

    \vspace{0.2cm} 

    \subfloat[\scriptsize THW (s)]{%
        \includegraphics[width=0.235\textwidth]{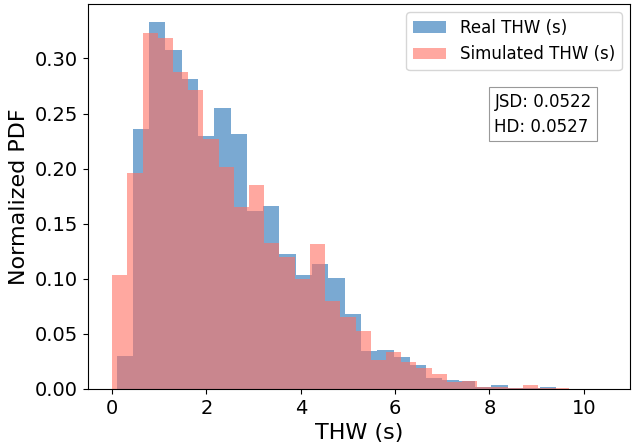}%
        \label{fig:thw3}%
    }
    \hspace{0.0\textwidth}
    \subfloat[\scriptsize iTTC (1/s)]{%
        \includegraphics[width=0.24\textwidth]{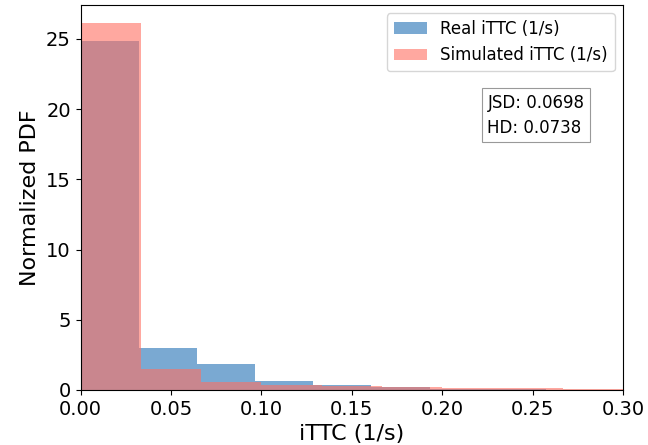}%
        \label{fig:iTTC4}%
    }

    \caption{Key distribution metrics to evaluate the generation of human-like traffic scenarios.}
    \label{fig:human-likeness}
\end{figure}

\textbf{Q2-Human-likeness} \cref{fig:human-likeness} shows the distributions of the simulation results and the real human driving data across four key indicators. Overall, the differences in the distributions for three of these indicators (minimum distance in \cref{fig:distance2}, THW in \cref{fig:thw3}, and iTTC in \cref{fig:iTTC4}) are very small, with both the JSD and HD values below 0.1. This indicates that our proposed method can accurately reproduce human driving behavior, particularly in terms of maintaining safe distances and avoiding collisions. It is worth noting that the simulated data shows a higher probability than the real data in the region where THW approaches 0, and several large outliers are observed in the iTTC distribution. These phenomena suggest that the learned policy may occasionally lead to a few high-risk or collision situations. 

The speed distribution in the simulation environment shows a certain degree of deviation compared to the real dataset, with JSD and HD values of 0.3530 and 0.3665, respectively. Unlike other safety indicators that exhibit minimal fluctuations, vehicles in highway scenarios typically operate within a wide range of speeds, making this level of deviation acceptable. As shown in \cref{fig:speed1}, the peak vehicle speed in the simulation is approximately 40 m/s, slightly higher than the 35 m/s observed in the real data. Despite the overall higher vehicle speeds, the simulated vehicles maintain reasonable and safe driving behavior, closely matching the real data in terms of key safety indicators.

Consequently, despite the differences in speed distribution and the presence of a small number of high-risk behaviors, our proposed method effectively simulates human-like driving behaviors by utilizing the reconstructed underlying reward function that corresponds to human driving data for each driving style. The simulation closely matches human driving behavior in terms of key safety metrics.

\begin{table*}[ht]
\centering
\caption{Evaluation of different methods using goal-reaching, off-road and collision rates}
\renewcommand{\arraystretch}{1.2}
\setlength{\tabcolsep}{12pt}
\begin{tabular}{llccc}
\toprule
\textbf{Method} & \textbf{Policy} & \textbf{Goal-reaching Rate (\%)} & \textbf{Off-road Rate (\%)} & \textbf{Collision Rate (\%)} \\
\midrule
   & Aggressive &35.31 &49.21 &15.42 \\
\hspace{0.3cm}BC & Cautious   &26.62 &64.57 &8.76 \\
   & Normal     &26.21 &62.57 &11.17 \\
\midrule
 & Aggressive &0.00 &99.95 &0.00 \\
\hspace{0.1cm}TD3+BC & Cautious   &0.00 &97.42 &2.53 \\
       & Normal     &0.96 &98.18 &0.81 \\
\midrule
     & Aggressive &54.75 &27.40 &17.79 \\
\hspace{-0.1cm}SAC (Log-replay)& Cautious   &53.33 &39.35 &7.22 \\
    & Normal     &62.96 &28.05 &8.94 \\
\midrule
 & Aggressive &53.81 &30.54 &15.65 \\
\hspace{-0.3cm}\textbf{HAD-Gen (Log-replay)} & Cautious   &63.77 &21.95 &14.23 \\
                       & Normal     &63.80 &22.81 &13.34 \\
\midrule
 & Aggressive &\textbf{76.64} &\textbf{9.70} &\textbf{13.61} \\
\hspace{-0.3cm}\textbf{HAD-Gen (Self-replay)} & Cautious   &\textbf{85.08} &\textbf{11.62} &\textbf{3.22} \\
                         & Normal     &\textbf{90.96} &\textbf{2.08} &\textbf{6.91} \\
\bottomrule
\end{tabular}
\label{tab:results}
\end{table*}

\textbf{Q3-Generalibility} As shown in \cref{tab:results}, compared with the baselines, the three polices obtained through HAD-Gen based multi-agent training exhibit the highest overall performance. Notably, the normal policy achieves the best results, with a goal-reaching rate of 90.96\%, an off-road rate of only 2.08\%, and a collision rate of 6.91\%. These results indicate that our proposed method possesses strong generalization ability and can be effectively adapted to various highway scenarios. In contrast, the success rate of the three polices based on BC is only around 30\%. This is mainly because the BC method is prone to cumulative errors when dealing with out-of-distribution data, which leads to poor generalization performance. TD3+BC yields the poorest performance, making it nearly impossible for the agent to reach its goal as it did in the dataset. This is primarily because offline RL lacks real-time interaction with the environment, which prevents it from effectively acquiring key behaviors such as collision avoidance and staying on the road. Although TD3+BC utilizes the reward function derived from the weights trained through IRL, the reward function from the naturalistic driving dataset does not include penalties for off-road behavior and collisions, contributing to its poor performance. 

Therefore, we further use SAC, an online RL algorithm, as a baseline. SAC learns by interacting with the environment rather than from a dataset. Its highest success rate is 62.96\%, and the lowest collision and off-road rates are 8.94\% and 28.05\%. However, if we pre-train with TD3+BC, followed by training with the same SAC, no obvious improvement is shown, indicating that offline RL cannot help too much when focusing on collision or off-road avoidance. 

The SAC trained above is based on a single-agent training framework, which is the \textit{Log-replay} training strategy defined below. However, we can also employ a multi-agent training framework, which is named \textit{Self-replay} in our context:

\textit{Log-replay}: We train the policy by selecting one agent (ego-agent) while allowing the other agents (non-ego) in the scene to follow the trajectories from the dataset. Additionally, we use IDM to control the non-ego agents when their distances to the ego-agent fall below the threshold set by IDM, ensuring they behave as reactive non-ego agents.

\textit{Self-replay}: We employ a MARL training framework, namely centralized training and decentralized execution (CTDE), to train a policy using all agents' data in a scenario.  

Compared to the long-replay method, the self-replay-based HAD-Gen demonstrates substantial improvements across all indicators. Taking the normal policy as an example, the goal-reaching rate increased from 63.80\% to 90.96\%, representing an improvement of approximately 42.5\%. Meanwhile, the off-road rate and collision rate decreased from 22.81\% and 13.34\% to 2.08\% and 6.91\%, respectively. The other two policies also achieve similar performance improvements. Based on this study, we further demonstrate the superior performance of the HAD-Gen (Self-replay) method proposed in this paper, concluding that RL with multi-agents can achieve higher learning efficiency in realistic traffic scenarios.

\section{Conclusion and future work}\label{sec:conclusion}
In this paper, we propose HAD-Gen, a novel framework for AV scenario generation, capable of producing diverse, human-like driving scenarios with effective generalization ability. Specifically, HAD-Gen first classifies driver behaviors based on the risk level ratio features observed throughout the agents' lifetime; then reconstructs the reward functions using MaxEntIRL; and finally obtains three driving policies through a combination of offline RL and MARL. Based on the experiments, we found that: 1) Offline RL struggles to train a robust policy due to imbalanced data, particularly the absence of near-miss and collision data, which are crucial for learning safe and effective driving behaviors; 2) It is recommended to consider different underlying reward functions for different driving styles as drivers exhibit distinct driving preferences; 3) Self-replay is more helpful for training a plausible driving policy than log-replay.

In future work, we plan to extend our method to different map layouts and explore Generative Adversarial Imitation Learning (GAIL) \cite{bhattacharyya2022modeling}, which removes the need for explicit reward function design. Additionally, exploring situation-adaptive reward functions \cite{kwon2024adaptive} is a promising direction, as drivers may exhibit different preferences depending on the driving context. Moreover, toward developing safe and explainable AI for trustworthy autonomous driving \cite{kuznietsov2024avreview}, we are interested in developing interpretable representations of the learned driving policies so that their driving decisions can be explained and communicated effectively.

\section*{Acknowledgment}
The authors would like to thank Balint Gyevnar and Xiaoxing Lyu for their valuable discussions.

\bibliographystyle{IEEEtran}
\bibliography{references}

\end{document}